\documentclass[11pt]{article}

\PassOptionsToPackage{table}{xcolor}
\usepackage[final]{acl}

\usepackage{times}
\usepackage{latexsym}
\usepackage[T1]{fontenc}
\usepackage[utf8]{inputenc}
\usepackage{microtype}
\usepackage{float}
\usepackage{amsmath}
\usepackage{graphicx}
\usepackage{subcaption}
\usepackage{enumitem}
\usepackage[most]{tcolorbox} 

\usepackage{pifont}
\usepackage{booktabs}
\usepackage{makecell}
\usepackage{multirow}
\usepackage{afterpage}
\usepackage{listings}

\lstdefinestyle{trace}{
    basicstyle=\ttfamily\scriptsize,
    breaklines=true,
    breakatwhitespace=false,
    breakindent=0pt,
    columns=fullflexible,
    keepspaces=true,
    xleftmargin=8pt,
    xrightmargin=0pt,
    aboveskip=4pt,
    belowskip=4pt,
    frame=L,
    framerule=0.4pt,
    rulecolor=\color{gray!60},
    showstringspaces=false,
    upquote=true,
    escapeinside={(*@}{@*)},
}

\title{
 \vspace{-20pt}
  \makebox[\linewidth][l]{\includegraphics[height=0.75cm]{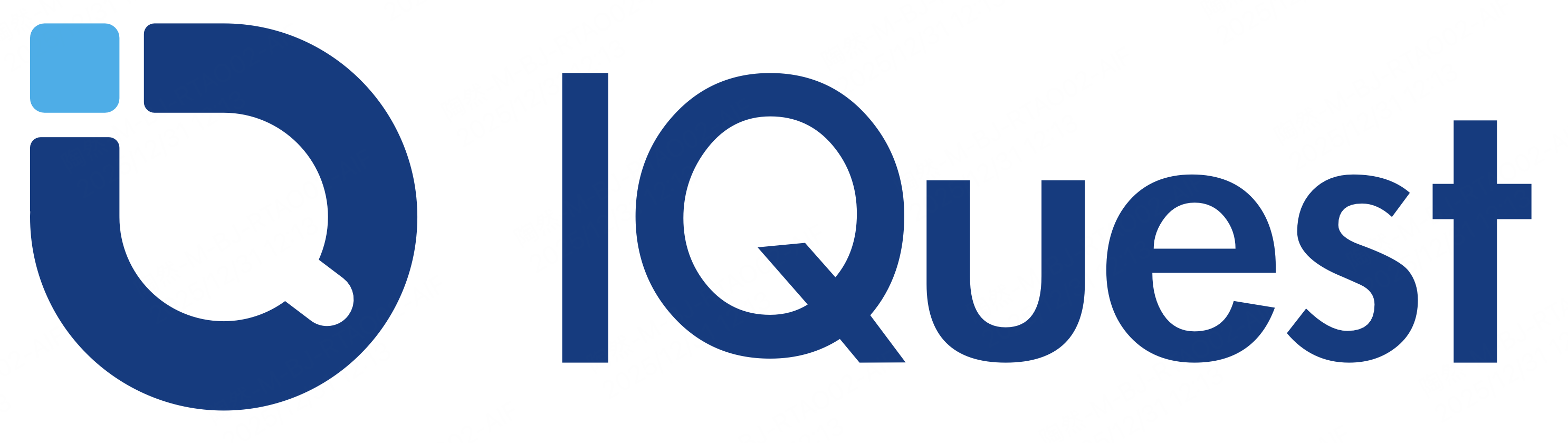}}\\[-10pt]
  \rule{\linewidth}{0.1pt}\\[5pt]
  MIRA: Mid-training Rubric Anchoring for Source-Aware Data Selection
}

\author{
  \textbf{Haowen Wang}$^{1}$\thanks{\ Equal contribution.}, 
  \textbf{Yaxin Du}$^{2,3}$\footnotemark[\value{footnote}], 
  \textbf{Jian Yang}$^{1}$\thanks{\ Corresponding authors.},
  \textbf{Jiajun Wu}$^{1,2}$ \\
  \textbf{Shukai Liu}$^{1,2}$\textbf{,}
  \textbf{Yuxuan Zhang}$^{2,4}$\textbf{,} 
  \textbf{Pingjie Wang}$^{3}$\textbf{,}
  \textbf{Siheng Chen}$^{3}$\\
  \textbf{Tuney Zheng}$^{2}$\thanks{\ Project leader.}\textbf{,}
  \textbf{Ming Zhou}$^{5}$\textbf{,}  
  \textbf{Xianglong Liu}$^{1}$\textbf{,}
  \textbf{Bryan Dai}$^{2}$\\[0.5em]
  $^{1}$Beihang University,
  $^{2}$IQuest Research \\
  $^{3}$Shanghai Jiao Tong University,
  $^{4}$University of British Columbia,
  $^{5}$Langboat
  \\[0.4em]
   \normalfont\includegraphics[height=1em]{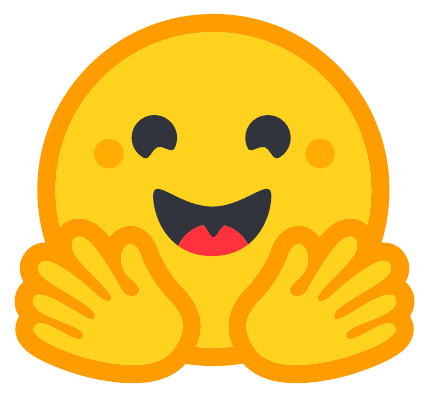}\;\href{https://huggingface.co/collections/Multilingual-Multimodal-NLP/mira}{\texttt{Multilingual-Multimodal-NLP/mira}}
}

\begin{document}
\maketitle

\begin{abstract}
Mid-training has become an important stage in modern LLM development, using large-scale curated mixtures to strengthen capabilities before final post-training. 
Its data selection problem is distinct: the data are optimized under a pretraining-style objective at near-pretraining scale, but are curated toward downstream capabilities and drawn from heterogeneous sources with different formats and training roles. 
As a result, effective selection requires both scalability and source-adaptive semantic criteria. 
Existing model-based methods scale well, but provide only implicit quality signals. Semantic selection methods offer stronger judgments, but usually assume fixed rubrics or standardized data formats.
To address this mismatch, we propose \textbf{MIRA}, a source-aware filtering framework based on \emph{self-anchored rubric discovery}.
The key idea is to make rubric construction part of data selection: MIRA first discovers what should be evaluated for each source group, then distills those judgments into scalable student scorers for full-corpus filtering. 
On code-oriented mid-training with 21 sources and 5 source groups, MIRA outperforms selection baselines across nine code benchmarks and matches the full-corpus run while using only half the tokens.
\end{abstract}


\section{Introduction}
\label{sec:intro}


Mid-training has become a key stage in modern large language model (LLM)~\citep{touvron2023llama,openai2023gpt4,yang2024qwen2,yang2025qwen3,yang2025codefoundationmodelsagents,yang2026incoder} development, but its data distribution differs from both broad pretraining and final post-training~\cite{liu2026midtrainingbridgespretrainingposttraining}. It keeps a pretraining-style objective and near-pretraining scale, while using curated data to strengthen downstream capabilities such as reasoning, coding, long-context understanding, and tool use~\cite{faircodegenteam2025cwmopenweightsllmresearch}. This creates a hybrid mixture: it includes pretraining-like sources such as filtered web documents, code, mathematics, and long technical text, as well as more structured sources such as instruction-style examples, reasoning traces, tool-use logs, and multi-turn agent trajectories~\citep{grattafiori2024llama3herdmodels,Guo_2025,wang2025octothinker,tu2025survey}. 
As a result, mid-training data form a large-scale, capability-oriented, and highly heterogeneous mixture of sources with different formats and training roles.

This combination makes it difficult to reuse data selection methods designed for either pretraining or post-training. 
Pretraining-oriented methods provide scalable signals such as perplexity, gradient information, influence estimates, or distributional matching~\citep{marion2023when,penedo2024fineweb,soldaini2024dolma,xie2023dsir,less2024,ntkselector2025}. 
However, these methods rely on implicit model- or distribution-level signals, and do not explicitly define the semantic criteria that make an example useful within its source format. 
For structured mid-training sources, a difficult or influential example may reflect useful signal, but it may also reflect formatting artifacts, inconsistent trajectories, or incorrect reasoning.
Post-training-oriented methods, including human-defined criteria, preference signals, LLM judgments, or learned quality scorers~\citep{du2023mods,ivison2025largescale,li2023one,wettig2024quratingselectinghighqualitydata,peng2025datamandatamanagerpretraining}, offer stronger semantic supervision, but typically assume fixed criteria or more standardized data formats. 
This makes them difficult to apply directly to mid-training mixtures, where sources differ in format, training role, and the evidence that indicates quality.
Thus, existing methods leave a mismatch: pretraining filters scale but do not define source-specific semantic quality, while post-training filters provide semantic supervision but assume the criteria are already known.

To address this gap, we propose \textbf{MIRA} (\textbf{Mi}d-training \textbf{R}ubric \textbf{A}nchoring for Source-Aware Data Selection), a group-wise quality scoring and filtering framework for heterogeneous mid-training corpora. 
The key idea is to separate \emph{rubric discovery} from \emph{scalable scoring}: MIRA uses a frontier teacher only to determine what should be evaluated for each source group, and then distills the resulting judgments into lightweight student scorers for full-corpus filtering. 
Instead of applying a fixed global rubric to all data sources, MIRA constructs rubrics during data selection. This design gives MIRA three advantages. 
First, it adapts the quality space to each source group rather than forcing all data into a fixed global rubric. 
Second, it preserves semantic supervision while making scoring scalable through student distillation. 
Third, it supports calibrated filtering by aggregating scores with source-conditioned reliability estimates and applying source-aware retention thresholds.

Concretely, MIRA proceeds in four steps. 
First, it groups related sources by content-embedding similarity, so that each group contains records with similar capability roles and quality patterns. 
Second, it performs self-anchored rubric discovery: the teacher freely articulates quality dimensions for sampled records, and MIRA clusters these dimension-level judgments into \emph{group-specific anchor rubrics}. 
Third, these anchors are used to obtain structured teacher labels, which are distilled into group-specific student scorers for full-corpus inference. 
Fourth, MIRA applies source-conditioned reliability aggregation and source-aware retention thresholds, masking unreliable dimensions and avoiding a single global cutoff. 
Together, these steps make data selection source-adaptive at the rubric level, scalable at the scoring level, and calibrated at the mixture level.

For empirical validation, we use code-oriented mid-training as the downstream setting. As code is a canonical capability domain with heterogeneous sources, including code documents, code-generation QA pairs, and agent tool-use records, generic language evaluations would tend to average away the source-level quality differences that MIRA is designed to capture~\citep{yang2026iquestcoderv1technicalreport,yang2025codefoundationmodelsagents}. 
In this setting, spanning 21 sources and 5 source groups, MIRA uses over 2M teacher-scored records for rubric discovery and distillation and applies the resulting student scorers to tens of millions of records. When aggregated over nine benchmarks, its group-level variant outperforms strong 50B-token  baselines and slightly exceeds the unfiltered full-corpus run while using half the tokens.

Our contributions are threefold:
\begin{itemize}[topsep=0.1em,itemsep=0.01em,leftmargin=1em]
    \item We study heterogeneous mid-training data selection from the perspective of \emph{source-adaptive semantic quality assessment}. 
    Instead of relying on a single global score, we ask how quality criteria can be derived and applied across diverse capability-oriented sources.

    \item We introduce \textbf{MIRA}, a rubric anchoring framework that derives group-specific quality criteria from sampled records. 
    MIRA groups related sources and uses a frontier teacher model to discover source-relevant semantic dimensions, avoiding fixed global rubrics.

    \item We develop a scalable source-aware filtering pipeline that converts discovered rubrics into structured teacher labels, distills them into group-specific student scorers, and applies reliability-aware aggregation with source-specific retention thresholds for full-corpus selection.
\end{itemize}

\section{Related Work}
\label{sec:related}

\subsection{Mid-training}

Mid-training has emerged as a dedicated stage between large-scale pretraining and task-specific post-training, designed to strengthen capabilities that are underrepresented or weakly organized in general web corpora~\citep{tu2025survey}. 
Unlike domain-specific continued pretraining, which primarily adapts a model toward a narrow target distribution~\citep{zhang2025adept}, mid-training works at a broader capability level, improving skills such as coding, mathematical reasoning, and long-context understanding while preserving general language competence. 
Empirical studies show that mid-training is especially effective when the pretraining and post-training distributions diverge, such as in code and mathematics~\citep{liu2026midtrainingbridgespretrainingposttraining}, and can further amplify the gains of subsequent reinforcement learning~\citep{wang2025octothinker}. 
A complementary line of work studies how to mix and schedule domains during continued training, including dynamic sampling based on learning velocity~\citep{luo2024velocitune} and scaling laws for domain mixture ratios~\citep{gu2024cmr}. 
These works establish the importance of mid-training data composition, but they mainly focus on stages, domains, mixtures, or schedules, whereas MIRA focuses on the finer-grained question of how to select individual records from heterogeneous mid-training sources.

\subsection{Training Data Selection}

Training data selection methods can be broadly grouped into three categories. 
(1) Heuristic and corpus-level filtering methods use deduplication, perplexity, toxicity filtering, educational-value scoring, domain classification, or lightweight classifiers to curate large web corpora~\citep{thrush2025improvingpretrainingdatausing,soldaini2024dolma,penedo2024fineweb,li2025datacomplmsearchgenerationtraining,wang2025ultrafinewebefficientdatafiltering}. 
These methods are scalable, but provide limited semantic evidence about whether a structured example teaches the intended capability. 
(2) Optimization- and target-aware selection methods use proxy losses, gradients, influence estimates,  or distribution matching to select data for a target distribution or evaluation objective~\citep{xie2023dsir,shum2025predictive,engstrom2024dsdm,less2024,ntkselector2025,pan2024gdig,du2025feddqc,zhao2024enhancing}. 
They are more directly tied to downstream utility, but their signals remain implicit. 
(3) Semantic and rubric-based selection methods use human-defined criteria, preference signals, LLM-as-judge annotations, or learned scorers to evaluate instruction quality, diversity, educational value, or generic usefulness~\citep{du2023mods,li2023one,ivison2025largescale,wu2024best,wang2024selftaught,wettig2024quratingselectinghighqualitydata,peng2025datamandatamanagerpretraining}. 
These methods are more interpretable, but typically rely on fixed or globally shared criteria. 
MIRA differs by making rubric construction source-adaptive: it induces group-specific anchor rubrics from sampled records, distills them into student scorers, and applies source-aware filtering across the full corpus.


\section{Method}
\label{sec:method}

\begin{figure*}[!t]
    \centering
    \includegraphics[width=\linewidth]{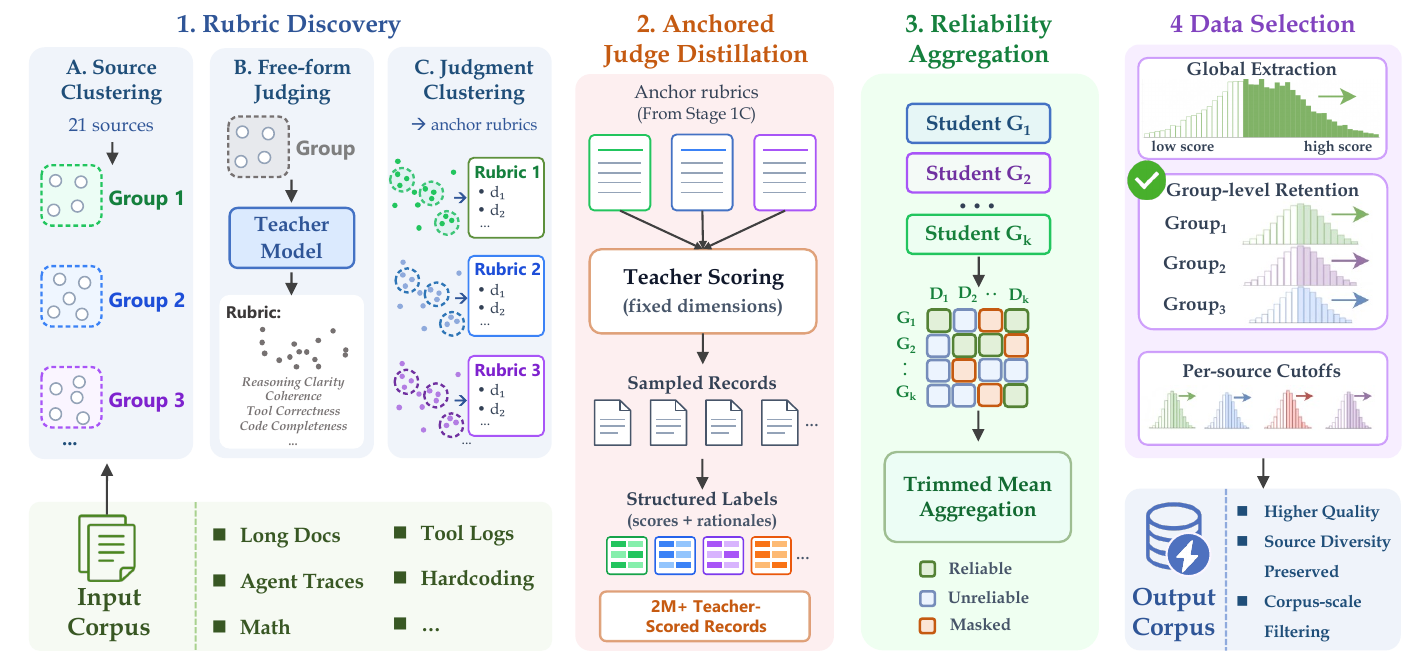}
    \vspace{-0.6cm}
    \caption{
        Overview of the \textbf{MIRA} pipeline.
        Heterogeneous mid-training sources are first organized into capability-coherent
        groups. Within each group, a frontier judge produces free-form judgments that are
        clustered into a fixed set of \emph{anchor dimensions}
        (\S\ref{sec:rubric-discovery}).
        These anchors define a stable scoring space in which the teacher re-scores a
        larger sample, and the resulting structured labels are distilled into a
        group-specific student scorer that runs over the full corpus
        (\S\ref{sec:distillation}).
        Per-record student outputs are then aggregated under a source-conditioned
        reliability mask that suppresses unreliable source--dimension pairs
        (\S\ref{sec:reliability}), and per-source retention thresholds convert the
        aggregated scores into the final selection while preserving source diversity
        (\S\ref{sec:selection}).
    }
    \vspace{-0.3cm}
    \label{fig:pipeline}
\end{figure*}

\subsection{Overview}
\label{sec:overview}

MIRA addresses heterogeneous mid-training data selection by making rubric construction part of the filtering pipeline. The key challenge is that different sources expose different capability signals, so the quality criteria cannot be assumed to be fixed or globally shared. 
MIRA therefore first discovers source-group-specific rubrics from sampled records, then distills the judgments into efficient scorers that can be applied to the full corpus.

As shown in Figure~\ref{fig:pipeline}, MIRA consists of four modules:  \emph{self-anchored rubric discovery} (\S\ref{sec:rubric-discovery}),  \emph{anchored judge distillation} (\S\ref{sec:distillation}),  \emph{source-conditioned reliability aggregation} (\S\ref{sec:reliability}), and \emph{source-preserving selection} (\S\ref{sec:selection}). The first module derives group-specific anchor rubrics. The second converts these anchors into distillable teacher labels and student scorers. The third aggregates student scores using source-conditioned reliability estimates. The fourth converts aggregated scores into a filtered corpus through source-aware retention. This decomposition separates the semantic part of selection from the scalable part: the frontier judge is used to identify what should be scored for each source group, while student scorers and source-aware aggregation perform full-corpus filtering.

\subsection{Self-Anchored Rubric Discovery}
\label{sec:rubric-discovery}

Mid-training sources each carry a distinct capability profile, so no single global rubric can faithfully evaluate all of them. Yet manually authoring a separate rubric per source is impractical at the scale of a modern mid-training corpus. MIRA therefore operates at the granularity of \emph{source groups}, automatically inducing group-specific rubrics from the teacher's own judgments.

\paragraph{Source Clustering and Free-form Judging.}
We first embed a representative sample of records from each source and cluster sources by the mean content embedding, forming a small set of capability-coherent groups. Within each group, a frontier judge is presented with sampled records and asked to freely propose quality dimensions, assign scores, and provide reasons, without any pre-specified rubric (see Appendix~\ref{app:prompt-phase1} and Figure~\ref{lst:prompt-freedim} for the full prompt template). This free-form protocol lets the teacher surface the quality concepts it actually uses, rather than confirming or denying criteria prescribed by the authors.

\paragraph{Judgment clustering and anchor extraction.}
Each free-form response is parsed into a set of \emph{judgment points}, where each point is a (dimension name, reason) pair. We embed all judgment points within a group and cluster them. For each cluster, we select the point nearest to the centroid as the \emph{anchor dimension} for that cluster. Across all clusters in a group, this yields a fixed set of anchor dimensions in our implementation, which compactly represents the teacher's quality vocabulary for that group.

Source clustering avoids the over-coarseness of a global rubric while remaining more tractable than per-source design. Free-form judging ensures that the anchors emerge from the teacher's actual assessment behavior rather than author intuition, making the discovered rubric an empirical artifact rather than a normative one.

\subsection{Anchored Judge Distillation}
\label{sec:distillation}

Free-form judgments are well-suited to rubric discovery, but because each sampled record receives a different set of dimensions, they cannot directly serve as training labels for a student scorer that must produce consistent output across all records. We therefore fix the discovered anchors and use them to construct a stable, distillable label space.

\paragraph{Anchored teacher scoring.}
For each source group, we re-prompt the frontier teacher with the group's anchor dimensions and ask it to score a larger teacher-labeling sample on every anchor, generating both a numerical score and a brief rationale per dimension (see Appendix~\ref{app:prompt-phase2} and Figure~\ref{lst:prompt-anchored} for the full prompt template). Because the scoring space is now fixed, records within a group are labeled in a comparable, structured way that supports direct supervision.
We split these anchored teacher-scored records into a training split for student distillation and a validation split for reliability diagnostics.

\paragraph{Student distillation.}
A frontier teacher is too expensive to run over tens of millions of mid-training records.
We therefore distill the anchored teacher labels into \emph{group-specific generative
student scorers}: lightweight models fine-tuned to produce, for each input record, a
score and rationale for every anchor dimension in the group's rubric.

We train a separate student per source group rather than a single universal scorer.
Because each group's anchors are internally coherent---reflecting a shared capability
theme---a group-specific student operates in a more homogeneous rubric space and can
fit the teacher more faithfully than a multi-group model would.
Generating scores together with rationales, rather than predicting a single scalar,
keeps student outputs interpretable, parseable into per-dimension values, and
compatible with the dimension-level reliability diagnostics in the next stage.

\subsection{Source-Conditioned Reliability Aggregation}
\label{sec:reliability}

Even a well-trained student scorer may not be uniformly reliable across all sources and
all dimensions.
A source that is rare in the distillation sample, or a dimension that is weakly
predictable from text features alone, can exhibit high teacher--student disagreement.
Averaging all dimensions without accounting for this variance would let unreliable
dimensions corrupt the overall score and distort downstream filtering decisions.

\paragraph{Residual diagnostics.}
For each source group $g$, let $\mathcal{S}_g$ denote the sources in the group and $\mathcal{D}_g$ denote its discovered anchor dimensions. For each source--dimension pair $(s,d)$ with $s \in \mathcal{S}_g$ and $d \in \mathcal{D}_g$, we evaluate the student on the validation split and compute teacher--student agreement statistics (MAE and Spearman correlation). Pairs that fall below reliability thresholds on either metric are flagged as unreliable. These statistics collectively form a \emph{source-conditioned reliability mask}
$M^{(g)} \in \{0,1\}^{|\mathcal{S}_g| \times |\mathcal{D}_g|}$ for each group.

\paragraph{Post-hoc masking and robust aggregation.}
Rather than modifying the student's prompt or re-running inference, we apply the reliability mask \emph{post-hoc} at aggregation time: for a record from source $s$, only the dimensions $d$ in its group $g$ with $M^{(g)}_{s,d}=1$ contribute to the overall score, which is computed as a trimmed mean over surviving dimensions (Figure~\ref{fig:radar_per_group}). Keeping the student prompt fixed is important for two reasons. First, removing a dimension from the prompt changes the joint distribution of the remaining dimensions, inducing prompt-driven score drift.
Second, post-hoc masking allows the diagnostic strategy to be updated at low cost, as a flagged dimension can be masked without rerunning full-corpus inference.

\subsection{Source-Preserving Selection}
\label{sec:selection}

Score distributions vary across mid-training sources: a source of concise
mathematical derivations and a source of long agent trajectories will have different mean scores and different spreads, reflecting capability differences rather than uniform quality. Applying a single global cutoff to all sources will therefore drain the lower-mean sources first, and because each source corresponds to a capability region, source attrition translates directly into capability attrition.

\paragraph{Selection granularity.}
We instantiate the final retention step at three granularities, corresponding
to the variants evaluated in Table~\ref{tab:main}. \textbf{MIRA-Global} applies
a single threshold over the entire scored corpus, selecting the highest-scoring
records regardless of source. This maximizes global score ranking but can
over-select groups whose score distributions are naturally higher. \textbf{MIRA-Group}
applies retention thresholds within each capability-coherent source group.
Because records in the same group share a discovered rubric and a group-specific
student scorer, their scores are better calibrated, and group-level retention
therefore preserves broad capability coverage while still allowing high-quality
sources within a group to compete. \textbf{MIRA-Source} applies thresholds
separately within each source, providing the strongest source-diversity
preservation but relying on finer-grained and potentially noisier source-level
score distributions.

Among these variants, MIRA-Group is intended as the default trade-off: it avoids the capability collapse of global filtering while remaining more stable than fully per-source thresholding for small or noisy sources.




\section{Experiments}
\label{sec:experiments}

\begin{table*}[t]
\small
\centering
\caption{
  End-to-end evaluation after mid-training and SFT across nine benchmarks and four categories.
  We compare the full 50B-token mid-training corpus with 25B-token selection methods.
  Multipl-E is averaged over eight languages, and Macro Avg. averages the four category scores.
  \textbf{Bold} marks the best score in each column.
}
\vspace{-0.3cm}
\label{tab:main}
\resizebox{\textwidth}{!}{
\begin{tabular}{l|cccccc|c|ccc|c|>{\columncolor[HTML]{EFEFEF}}c}
\toprule
\multirow{2.4}{*}{\bf Method}
& \multicolumn{6}{c|}{\bf Code Generation}
& \bf Multilingual
& \multicolumn{3}{c|}{\bf SQL (EX)}
& \bf SWE
&  \\
\cmidrule(lr){2-7} \cmidrule(lr){8-8} \cmidrule(lr){9-11} \cmidrule(lr){12-12}
& MBPP & MBPP+ & BCB-f & BCB-h & LCB & Avg.
& MultiplE
& Spider & BIRD & Avg.
& Avg.
& \multirow{-2.4}{*}{\makecell{\bf Macro\\\bf Avg.}} \\
\midrule
\multicolumn{13}{c}{\textit{\textbf{No Mid-training}}} \\
\midrule
Base Model
& 84.10 & 69.30 & 51.40 & 23.65 & 5.99 & 46.89
& 59.14
& 11.22 & 1.37 & 6.29
& 0.33
& 28.16 \\
~~+SFT
& 86.80 & 72.50 & \bf 56.84 & 32.43 & 20.96 & 53.91
& 72.57
& 79.98 & 48.50 & 64.24
& 3.67
& 48.60 \\
\midrule
\multicolumn{13}{c}{\textit{\textbf{Full Corpus (50B tokens, no quality filtering)}}} \\
\midrule
Raw Mixture
& 87.60 & 72.50 & 53.51 & 30.41 & 24.55 & 53.71
& 67.42
& 98.07 & 90.29 & 94.18
& \bf 40.00
& 63.83 \\
\midrule
\multicolumn{13}{c}{\textit{\textbf{25B-Token Subset Selection (half of the full corpus)}}} \\
\midrule
DSIR
& 86.80 & 72.20 & 40.18 & 22.97 & 21.56 & 48.74
& 67.26
& \bf 98.36 & \bf 92.05 & \bf 95.20
& 27.00
& 59.55 \\
PPL
& 83.60 & 70.10 & 52.37 & 26.76 & 19.76 & 50.52
& 57.74
& 90.99 & 90.33 & 90.66
& 20.00
& 54.73 \\
Random
& 86.80 & 70.90 & 55.00 & 25.68 & \bf 25.15 & 52.71
& 71.44
& 96.52 & 91.07 & 93.79
& 35.00
& 63.23 \\
DataMan
& 87.80 & 72.50 & 55.79 & 29.05 & 23.95 & 53.82
& 71.38
& 97.58 & 90.09 & 93.84
& 33.00
& 63.01 \\
\rowcolor[HTML]{E8E8FF}
\textbf{MIRA-Global~(Ours)}
& 88.40 & \bf 73.80 & 52.19 & 29.05 & 22.16 & 53.12
& 67.84
& 97.00 & 91.53 & 94.26
& 32.00
& 61.81 \\
\rowcolor[HTML]{E8E8FF}
\textbf{MIRA-Group~(Ours)}
& \bf 88.90 & \bf 73.80 & 55.26 & \bf 33.11 & 21.56 & \bf 54.53
& 71.85
& 98.26 & 89.90 & 94.08
& 36.33
& \bf 64.20 \\
\rowcolor[HTML]{E8E8FF}
\textbf{MIRA-Source~(Ours)}
& 87.30 & 72.20 & 55.09 & 31.76 & 24.55 & 54.18
& \bf 72.84
& 97.49 & 91.26 & 94.38
& 30.33
& 62.93 \\
\bottomrule
\end{tabular}}
\vspace{-5pt}
\end{table*}

\noindent \textbf{Setup.}
We use \textbf{Qwen2.5-Coder-14B}~\citep{hui2024qwen25coder} as the base model for all experiments.
Mid-training is conducted using Megatron-LM for approximately 50B tokens,
with sequence length 128k, global batch size 256, and BF16 precision.
After mid-training, each checkpoint is fine-tuned on a fixed set of 400K
instruction-following samples. All SFT hyperparameters are held constant across
conditions so that any performance difference reflects only the mid-training
data selection strategy. Full mid-training and SFT configurations,
including parallelism, checkpointing, and sequence-packing details, are
reported in Appendix~\ref{app:midtrain-config} and~\ref{app:sft-config}.

\noindent \textbf{Scoring stack.}
The MIRA pipeline of \S\ref{sec:method} is instantiated with
\textbf{Kimi-K2.6}~\citep{kimik26} as the frontier teacher for both Phase-1 free-dim
rubric discovery and Phase-2 anchored scoring, and with one
\textbf{Qwen3.5-35B-A3B-Base}~\citep{qwen3.5} student scorer per source group (a 35B-parameter mixture-of-experts decoder with $\approx 3$B
active parameters per token), full-parameter fine-tuned on the
teacher's anchored labels. Across the 5 source groups, the Phase-2
anchored corpus comprises approximately 2M teacher-scored records,
split into a training split for student distillation and a held-out
validation split for the reliability diagnostics of
\S\ref{sec:reliability}. Full teacher / student configurations,
per-group training recipe, and compute budget are reported in
Appendix~\ref{app:setup}.

\noindent \textbf{Baselines.}
We compare MIRA with no-mid-training controls (Base Model and SFT-only), a
full-corpus Raw Mixture, and four 25B-token selection baselines: source-preserving
Random sampling, perplexity filtering~\citep{marion2023when}, DSIR importance
resampling~\citep{xie2023dsir}, and the DataMan quality scorer~\citep{peng2025datamandatamanagerpretraining}.
All methods use the same raw corpus, and all filtered corpora are matched
in total token count. Detailed baseline definitions are provided in
Appendix~\ref{app:baseline}.

\noindent \textbf{Evaluation.}
We evaluate each model after the shared SFT stage on four benchmark groups.
For code generation, we report MBPP~\citep{DBLP:journals/corr/abs-2108-07732},
MBPP+~\citep{liu2023codegeneratedchatgptreally}, BigCodeBench-Full (BCB-f) and
hard split(BCB-h)~\citep{zhuo2025bigcodebenchbenchmarkingcodegeneration},
and LiveCodeBench (LCB)~\citep{jain2024livecodebenchholisticcontaminationfree}.
The group average is the macro-average over these five scores. For multilingual
code generation, we use Multipl-E~\citep{cassano2022multiplescalableextensibleapproach}
and report the macro-average over eight programming languages. For SQL, we evaluate
executable accuracy on Spider~\citep{DBLP:journals/corr/abs-1809-08887} and
BIRD~\citep{li2023llmservedatabaseinterface} and average the two scores. For
software-engineering, we report the SWE-Multi~\citep{jimenez2024swebench} pass rate.
The final macro average gives equal weight to the four groups: code generation,
Multipl-E, SQL, and SWE-M. 

\subsection{Main Results}
\label{sec:results}

Table~\ref{tab:main} reports post-SFT performance across four benchmark groups: code generation, multilingual code generation, executable SQL, and software-engineering repair. We compare the base model and its SFT-only variant against mid-trained models using either the full 50B-token corpus or 25B-token filtered subsets. All 50B-token methods are trained with the same token budget, so differences within the subset-selection block isolate the effect of the data selection criterion rather than extra compute.

\begin{figure}[!t]
    \centering
    \includegraphics[width=\linewidth]{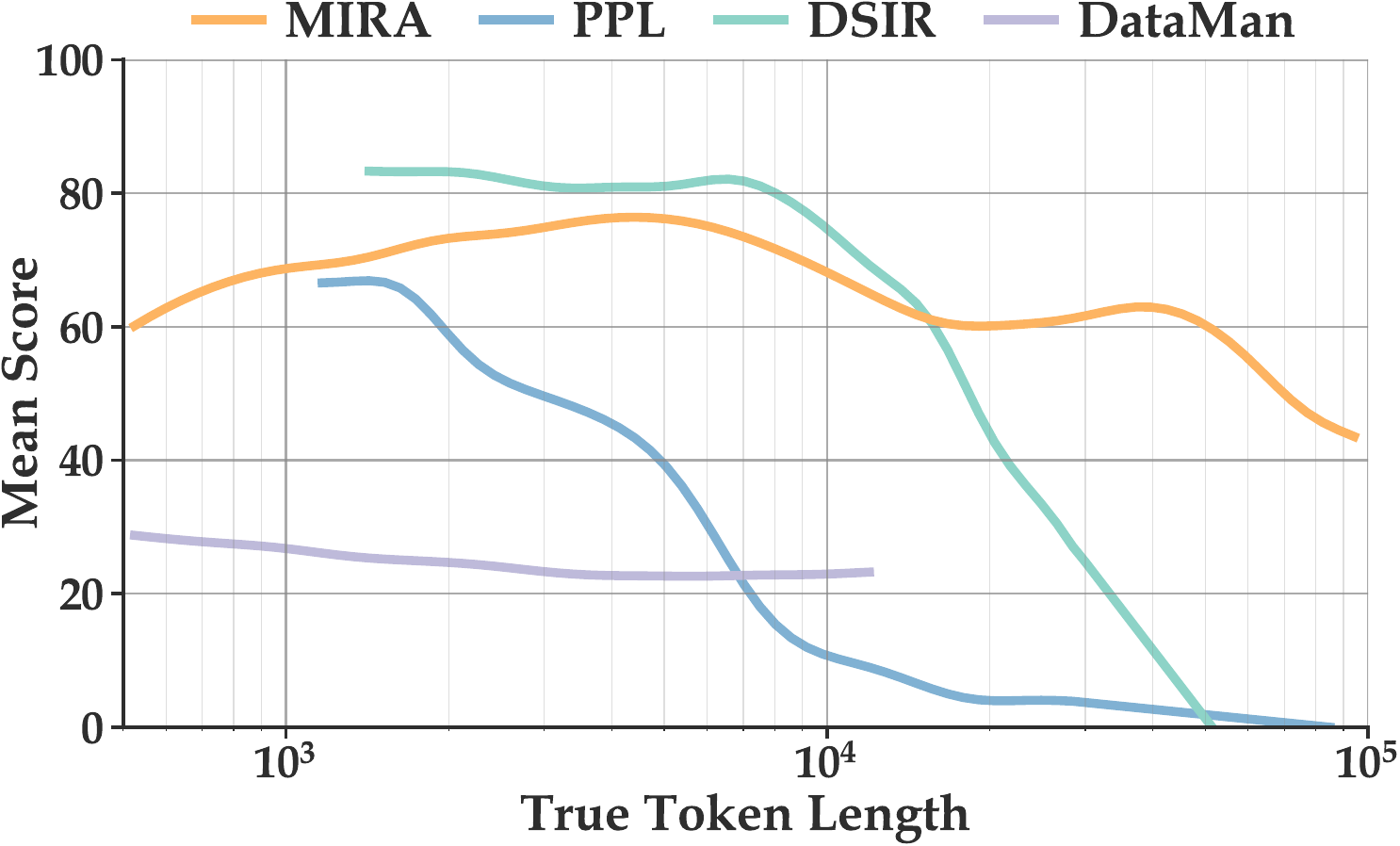}
    \vspace{-0.8cm}
    \caption{Statistics of length-conditioned scorer behavior. The $x$-axis shows token length on a log scale, and the $y$-axis reports the mean normalized score.}
    \label{fig:dist_per_group}
    \label{fig:combined_radar_dist}
    \vspace{-0.3cm}
\end{figure}

The results show three main trends. (1) MIRA provides the strongest overall 25B-token trade-off: MIRA-Group reaches the best macro average, 64.20, compared with 63.23 for Random, 63.01 for DataMan, and 59.55 for DSIR, while using only half of the full raw corpus. 
Random remains competitive because it preserves source diversity, while generic quality and distribution-matching baselines produce more imbalanced retained corpora (Figure~\ref{fig:dist_per_group}). (2) Different MIRA variants specialize in different capability groups. MIRA-Group obtains the best code-generation average, 54.53, with leading scores on MBPP, MBPP+, and BigCodeBench-hard at 88.90, 73.80, and 33.11, respectively. MIRA-Source achieves the best Multipl-E average, 72.84. (3) MIRA remains competitive on high-level task benchmarks where the best baseline varies. MIRA-Source reaches a 94.38 SQL average, close to the best DSIR score of 95.20, and MIRA-Group obtains 36.33 on SWE-Multi, exceeding all other 25B-token selection baselines.

\section{Analysis}
\label{sec:analysis}

\subsection{Scorer Analysis}
\label{sec:cross-format-distributions}
\begin{figure*}[t]
    \centering
    \includegraphics[width=\linewidth]{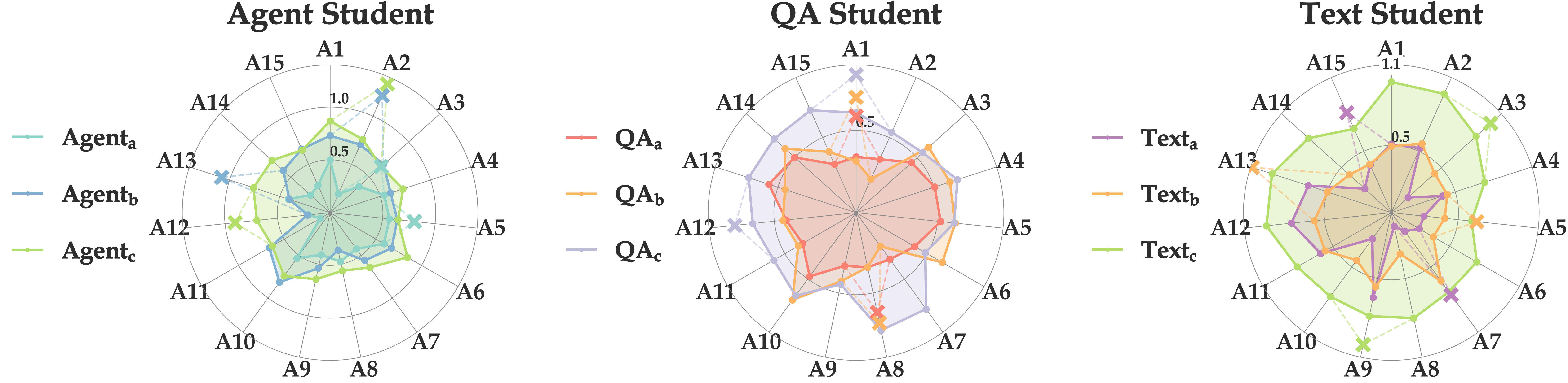}
    \caption{
        Reliability diagnostics for the Agent, QA, and Text student scorers. Each spoke \textbf{A1--A15} is one group-specific rubric dimension, and the radial value is teacher--student MAE on the validation split. The outlier $\times$ marks dimensions removed by the reliability mask before score aggregation.
    }
    \vspace{-0.3cm}
    \label{fig:radar_per_group}
\end{figure*}

To examine whether different scorers provide useful quality signals for heterogeneous training records, we analyze their behavior from the perspective of sequence length. 
We bucket records by token length, sample up to $1{,}000$ records from each $10$K-token interval, and score the records with PPL, DSIR, DataMan, and MIRA. 
Figure~\ref{fig:dist_per_group} plots normalized scores within each length bin, with token length shown on a log scale and the vertical line marking the mid-training length cutoff.

Figure~\ref{fig:dist_per_group} shows that: \textbf{(1) Baseline scorers exhibit strong length dependence.} PPL drops sharply as token length increases, while DSIR remains high on short records but collapses near the long-context region. Both trends suggest that their scores are partly shaped by sequence length rather than only by record quality. \textbf{(2) DataMan is less length-sensitive but has limited long-context coverage.} Its scores stay in a relatively narrow range over the short and medium-length bins, suggesting that it is not as directly length-dependent as PPL or DSIR. However, DataMan fails to return scores for very long records, because its scorer is not designed to process long structured traces beyond its input-length limit. As a result, the curve disappears in the long-context region, where many agent trajectories appear, making DataMan difficult to use as a selection signal for this part of the corpus. \textbf{(3) MIRA is length-robust.} MIRA maintains a smooth score profile across short and medium-length records and remains usable around the training cutoff, instead of collapsing as length increases.

\subsection{Reliability Masking}
\label{sec:reliability-diagnostics}

Next, we inspect the reliability of the distilled student scorers. For each of
the Agent, QA, and Text groups, we compute teacher-student MAE on the validation split for all 15 group-specific dimensions. Figure~\ref{fig:radar_per_group}
plots the per-dimension residual curves for sources within each group. Each
$\times$ marks a source--dimension cell removed by the post-hoc reliability mask
before record-level aggregation.

The radar diagnostics support the source-conditioned masking design. \textbf{(1) Student reliability is not uniform across groups.} The three panels show different residual profiles for Agent, QA, and Text students, even though all are evaluated over the same 15-anchor template within their group. This means a single global correction rule would either over-mask reliable regions or leave unreliable ones active. \textbf{(2) Unreliability is sparse and dimension-specific.} The correction markers concentrate on selected anchors rather than covering entire sources or entire students. For example, different markers appear around high-residual axes such as A2, A8, A12, and A13 depending on the group. \textbf{(3) Post-hoc masking preserves useful signal while removing unstable dimensions.} By applying the mask only to flagged source-dimension cells, MIRA keeps the remaining anchor scores unchanged and
prevents a small number of high-MAE dimensions from dominating the trimmed mean.

\subsection{Rubric Space Visualization}
\label{sec:reliability-diagnostics}

Next, we inspect the reliability of the distilled student scorers. For each of
types (Agent/QA/Text), we compute teacher--student MAE on the validation
split for all 15 group-specific anchor dimensions. Figure~\ref{fig:radar_per_group}
plots the per-dimension residual curves for sources within each group. Each
$\times$ marks a source--dimension cell removed by the post-hoc reliability mask
before record-level aggregation.

The radar diagnostics support the source-conditioned masking design.
\textbf{(1) Student reliability is not uniform across groups.} The three panels
show different residual profiles for Agent, QA, and Text students, even though
all are evaluated over the same 15-anchor template within their group. This
means a single global correction rule would either over-mask reliable regions or
leave unreliable ones active. \textbf{(2) Unreliability is sparse and
dimension-specific.} The correction markers concentrate on selected anchors
rather than covering entire sources or entire students. For example, different
markers appear around high-residual axes such as A2, A8, A12, and A13 depending
on the group. \textbf{(3) Post-hoc masking preserves useful signal while
removing unstable dimensions.} By applying the mask only to flagged
source--dimension cells, MIRA keeps the remaining anchor scores unchanged and
prevents a small number of high-MAE dimensions from dominating the trimmed mean.

\vspace{-0.1cm}
\subsection{Rubric Space}

\label{sec:rubric-geometry}

\begin{figure}[!t]
    \centering
    \begin{subfigure}[t]{0.49\linewidth}
        \centering
        \includegraphics[width=\linewidth]{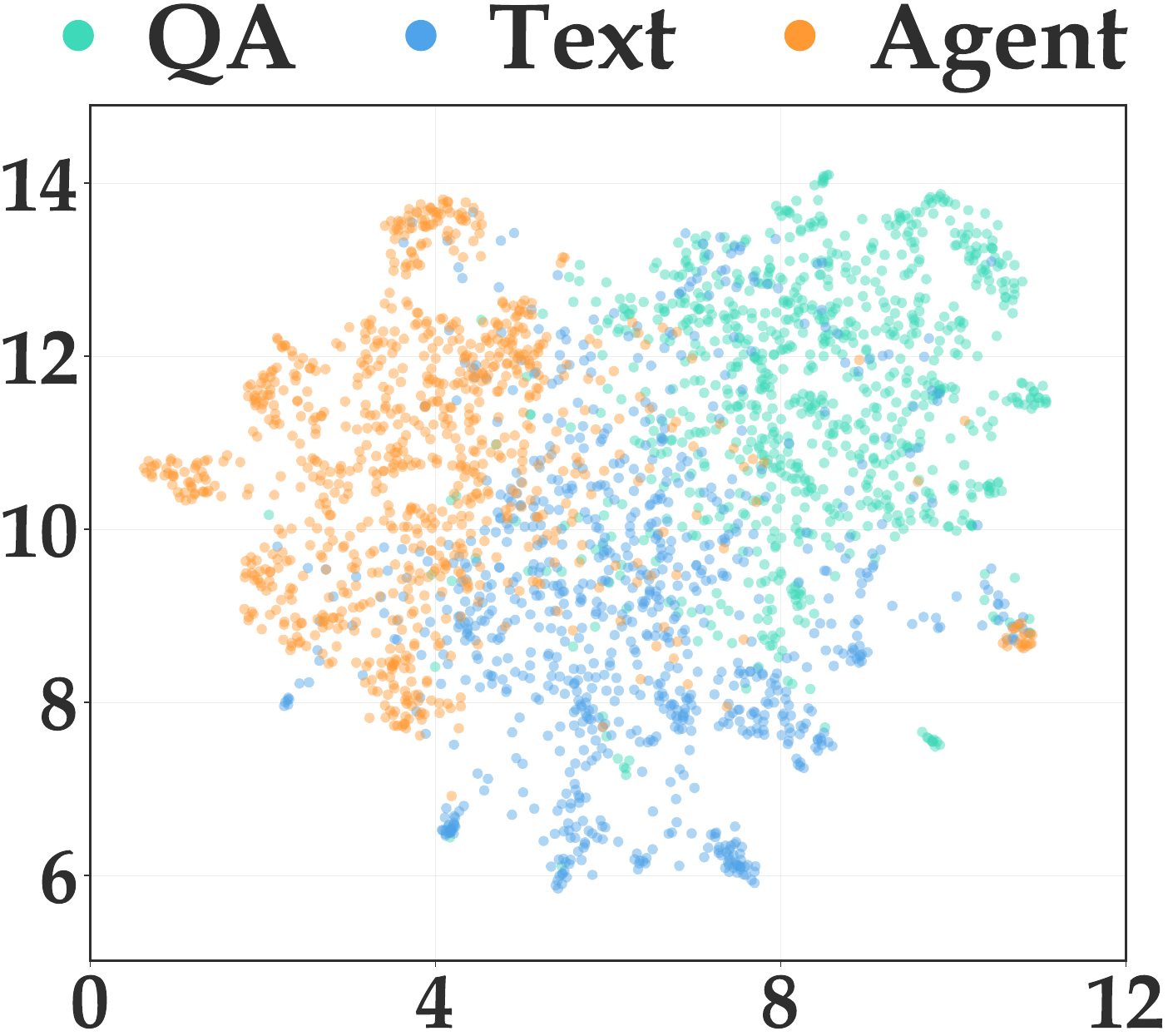}
        \vspace{-0.5cm}
        \caption{MIRA rubric t-SNE.}
        \label{fig:format_split}
    \end{subfigure}
    \hfill
    \begin{subfigure}[t]{0.49\linewidth}
        \centering
        \includegraphics[width=\linewidth]{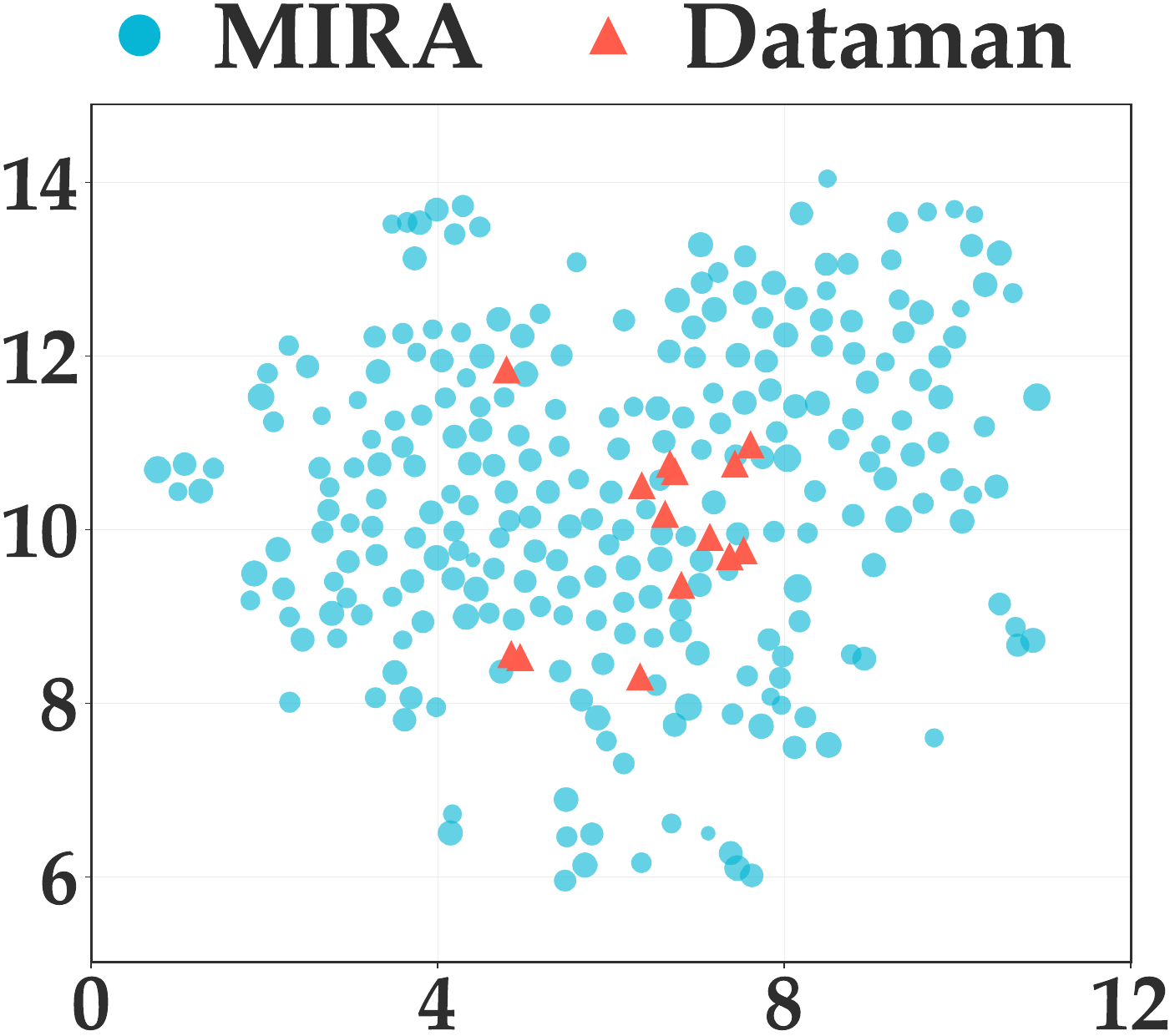}
        \vspace{-0.5cm}
        \caption{MIRA vs.\ DataMan t-SNE.}
        \label{fig:coverage}
    \end{subfigure}
    \vspace{-5pt}
    \caption{Visualization of MIRA rubic embeddings for QA / Text / Agent and the 14 DataMan rubics projected into the same dimension. \textbf{(a)} The three formats occupy mutually distinct regions of MIRA's rubic space.
    \textbf{(b)} All 14 DataMan rubics fall inside those regions.}
    \vspace{-0.4cm}
    \label{fig:rubric_geometry}
\end{figure}

To test whether MIRA discovers source-specific rubrics while still covering general-purpose quality criteria, we then compare the distribution of MIRA-discovered rubrics with the generic DataMan rubric. We embed MIRA rubric dimensions from 21 data sources, sample $3{,}000$ rubric points per format group, yielding $9{,}000$ QA, Text, and Agent points in total, and visualize it in a 2D t-SNE. We also embed the 14 DataMan rubric dimensions with the same encoder and project them into the same frame, allowing direct comparison with the final rubric dimensions discovered by MIRA.

The rubric distribution shows two properties. \textbf{(1) MIRA rubrics are
format-dependent.} In Figure~\ref{fig:format_split}, QA, Text, and Agent
rubrics occupy distinct regions, showing that different source formats induce
different quality criteria. This supports treating rubrics separately rather
than collapsing all sources into one global rubric. \textbf{(2) MIRA covers
DataMan while being more diverse.} In Figure~\ref{fig:coverage}, the 14 DataMan
rubric dimensions lie inside the broader MIRA rubric space. Quantitatively,
their nearest-neighbor distances to MIRA rubrics range from $0.160$ to $0.449$,
with $13/14$ inside MIRA's $P_{95}$ nearest-neighbor distance of $0.426$ and
$5/14$ below MIRA's median distance of $0.235$. Thus, MIRA retains the generic
quality dimensions represented by DataMan while expanding them into a more
diverse, source-aware rubric space.

\vspace{-0.1cm}
\subsection{Case Study}
\label{sec:agent-case-study}

\begin{figure}
    \centering
    \includegraphics[width=\linewidth]{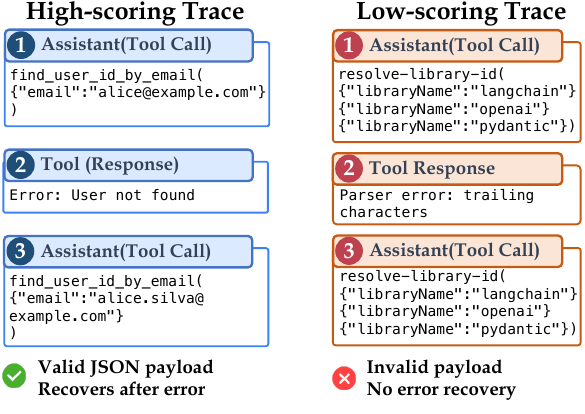}
    \vspace{-0.8cm}
    \caption{Case study of high- and low-scoring Agent traces. MIRA assigns low scores to traces with invalid tool-call payloads and no error recovery, rather than to superficially fluent but structurally correct traces.}
    \vspace{-0.3cm}
    \label{fig:agent-case-study}
\end{figure}
Finally, we inspect matched top-decile and bottom-decile traces from Agent sources to understand what low MIRA scores capture. 
As shown in Figure~\ref{fig:agent-case-study}, the high-scoring trace issues a valid tool-call payload, observes an error, and corrects the next action. 
The low-scoring trace instead concatenates multiple JSON objects into one \texttt{arguments} field, triggers a parser error, and repeats the same invalid call without recovery. 
Across inspected examples, the dominant low-score patterns are invalid tool-call payloads and lack of error reactivity.

This case study shows that MIRA's Agent scores reflect trajectory-level correctness rather than fluency alone. 
The good and bad traces are both fluent and multi-turn, but differ in whether tool calls are structurally valid and whether tool feedback changes subsequent actions. 
These source-specific failures would be difficult to capture with a single generic text-quality criterion, supporting the need for source-aware semantic filtering in heterogeneous mid-training data.

\section{Conclusion}

\label{sec:conclusion}

We presented \textbf{MIRA}, a source-aware filtering framework for heterogeneous mid-training data. MIRA discovers group-specific anchor rubrics from sampled records, distills them into efficient student scorers, and applies reliability-aware aggregation with source-aware retention thresholds for full-corpus selection. On 21 sources organized into 5 groups, MIRA-Group achieves the best 25B-token macro average across nine code-related benchmarks, outperforming PPL, DSIR, DataMan, and Random selection while matching the unfiltered 50B-token corpus at half the token budget. These results show that mid-training data selection benefits from source-adaptive quality criteria, not only scalable scoring.

\section*{Limitations}
MIRA addresses source-aware filtering for heterogeneous mid-training data by
deriving semantic quality scores beyond a single global criterion. However,
filtering is only one part of mid-training data management. Broader decisions
such as source discovery, mixture-ratio design, curriculum scheduling,
deduplication, and contamination control remain outside the scope of this work. Future work could include both integrating MIRA into broader data-mixture optimization pipelines and studying how source-aware quality scores interact with curriculum design and training-time sampling strategies.


\bibliography{custom}

\appendix
\clearpage
\section{Experimental Setup}
\label{app:setup}

\subsection{Baselines}
\label{app:baseline}

We compare MIRA against the following baselines:
\begin{itemize}[noitemsep,leftmargin=*]
  \item \textbf{Base Model}: The pretrained Qwen2.5-Coder-14B checkpoint without
        any mid-training, serving as a lower bound that shows the net gain of
        the mid-training stage itself.
  \item \textbf{Raw Mixture}: Mid-training on the full unfiltered corpus with no
        quality selection, establishing whether filtering provides any benefit
        over simply using all available data.
  \item \textbf{Random}: Within each source, records are
        randomly sampled to match the same per-source token budget as MIRA.
        This controls for the effects of training on fewer tokens and
        maintaining source distribution, isolating quality filtering from
        mere data reduction.
  \item \textbf{Perplexity Filtering}: Records are scored by the base model's
        per-token loss and filtered by perplexity thresholds, following the
        standard pretraining-style baseline~\citep{marion2023when}.
  \item \textbf{DSIR}: Data selection via importance resampling~\citep{xie2023dsir},
        which matches the mid-training corpus distribution to a high-quality
        target seed set using $n$-gram features.
  \item \textbf{DataMan}: A general-purpose quality scorer trained on
        multiple universal quality criteria~\citep{peng2025datamandatamanagerpretraining}, applied
        globally across all sources as a strong recent baseline for
        quality-based filtering.
\end{itemize}
All methods operate on the same raw corpus. Filtered corpora are matched in
total token count for a fair comparison.

\subsection{Data Sources and Grouping}
\label{app:data}

\paragraph{Sources and groups.}
Our mid-training experiments use 21 sources organized into 5
capability-coherent groups, each containing 3 to 5 sources:
two QA groups (\textbf{qa\_math\_reasoning} with 5 sources,
\textbf{qa\_code\_general} with 5), one Text group
(\textbf{text\_code\_doc} with 3), and two Agent groups
(\textbf{agent\_toolcall} with 5, \textbf{agent\_swe} with 3). Group
boundaries follow content-embedding similarity at the source level and
reflect a single capability theme per group: math/reasoning chains for
qa\_math\_reasoning, code and long-form technical QA for qa\_code\_general,
code-related long-form or extended-schema content for text\_code\_doc,
interactive tool-call traces for agent\_toolcall, and SWE repository
repair for agent\_swe. The 21 sources are drawn from a broader pool of
MIRA-scored records by dropping (i) HTML-frontend 
sources outside our code-focused evaluation, (ii) sources with high
content overlap with retained ones, and (iii) sources whose student
scorer (\S\ref{app:scorers}) is broadly miscalibrated.
Table~\ref{tab:data-sources} lists the per-source assignment, sampling
proportion, and final sampled record count.

\paragraph{Per-source sampling.}
Every source is independently down-sampled by a proportion
$p \in \{1.0, 0.5, 0.2, 0.1\}$ from a stage-2 corpus that was itself
stratified into low / mid / high quality bins by the group-specific
student scorer. Sources with $p{=}1.0$ retain stage-2's natural
${\approx}30 / 40 / 30$ low / mid / high distribution. Those with
$p{<}1.0$ are resampled so that the low / mid / high split becomes
$50 / 30 / 20$, deliberately keeping a long tail of mid- and
low-band records on heavy-volume sources so the corpus is not
over-concentrated on a single quality band. The 21-source totals after
this sampling come to ${\approx}11.6$M records
(${\approx}100$B raw tokens), and the 25B-token and 50B-token slices
of Table~\ref{tab:main} are drawn from this pool.

\begin{table}[!t]
\small
\centering
\caption{
  The 21 mid-training sources used in Table~\ref{tab:main}, organized
  into the 5 experiment-time groups of \S\ref{app:data}. ``Prop.'' is
  the per-source sampling proportion applied at stage-3 over the
  stage-2 low / mid / high bins. ``Rows'' is the resulting sampled
  record count. Per-group subtotals in italic.
}
\label{tab:data-sources}
\begin{tabular}{lrr}
\toprule
\textbf{Source} & \textbf{Prop.} & \textbf{Rows} \\
\midrule
\multicolumn{3}{l}{\textit{qa\_math\_reasoning} (5 sources)} \\
\midrule
math\_theorem                & $1.0$ & 416{,}397 \\
math\_code                & $1.0$ & 722{,}093 \\
math\_reason\_a                     & $0.1$ & 943{,}130 \\
math\_reason\_b          & $0.1$ & 279{,}689 \\
math\_reason\_c            & $0.5$ & 1{,}362{,}896 \\
\textit{Subtotal}       &       & \textit{3{,}724{,}205} \\
\midrule
\multicolumn{3}{l}{\textit{qa\_code\_general} (5 sources)} \\
\midrule
code\_qa\_a  & $0.1$ & 1{,}798{,}992 \\
code\_qa\_b     & $0.1$ & 611{,}274 \\
code\_qa\_c           & $0.1$ & 1{,}380{,}913 \\
code\_qa\_d             & $0.1$ & 125{,}631 \\
general\_qa            & $0.1$ & 542{,}672 \\
\textit{Subtotal}       &       & \textit{4{,}459{,}482} \\
\midrule
\multicolumn{3}{l}{\textit{text\_code\_doc} (3 sources)} \\
\midrule
algo\_text          & $0.2$ & 1{,}041{,}130 \\
pr\_text               & $1.0$ & 4{,}791 \\
code\_contest\_qa${}^{*}$    & $0.5$ & 126{,}850 \\
\textit{Subtotal}       &       & \textit{1{,}172{,}771} \\
\midrule
\multicolumn{3}{l}{\textit{agent\_toolcall} (5 sources)} \\
\midrule
cli\_agent\_a             & $0.5$ & 254{,}989 \\
cli\_agent\_b              & $0.5$ & 20{,}024 \\
funccall\_agent\_a         & $0.5$ & 121{,}702 \\
funccall\_agent\_b                & $1.0$ & 122{,}877 \\
funccall\_agent\_c                  & $1.0$ & 1{,}585{,}599 \\
\textit{Subtotal}       &       & \textit{2{,}105{,}191} \\
\midrule
\multicolumn{3}{l}{\textit{agent\_swe} (3 sources)} \\
\midrule
agent\_a        & $0.5$ & 27{,}547 \\
agent\_b             & $0.5$ & 36{,}941 \\
agent\_c                 & $1.0$ & 106{,}139 \\
\textit{Subtotal}       &       & \textit{170{,}627} \\
\midrule
\textbf{Total (21 sources)} &     & \textbf{11{,}632{,}276} \\
\bottomrule
\end{tabular}
\\[2pt]
{\footnotesize ${}^{*}$ code\_contest\_qa uses the qa-with-think 20-slot prompt variant. See \S\ref{app:scorers}.}
\vspace{-5pt}
\end{table}

\subsection{Mid-training Configuration}
\label{app:midtrain-config}

All mid-training runs in Table~\ref{tab:main} share the same Megatron-LM
launch script. Only the training-data path differs across the Random /
DSIR / PPL / DataMan / MIRA-Global / MIRA-Group / MIRA-Source / Raw-Mixture
conditions, so any post-SFT performance difference is attributable to the
mid-training data alone. Table~\ref{tab:midtrain-config} consolidates the
model architecture inherited from Qwen2.5-Coder-14B~\citep{hui2024qwen25coder}
together with the optimization, parallelism, and checkpointing settings.

\begin{table*}[!t]
\small
\centering
\caption{
  Mid-training configuration shared by every condition in Table~\ref{tab:main}.
  Architecture is inherited from Qwen2.5-Coder-14B and reproduced here for
  self-containment. TRAIN\_ITER is derived as
  $\lfloor \text{TOTAL\_TOKENS} / (\text{global-batch} \cdot \text{seq-len}) \rfloor$:
  the 50B-token Raw-Mixture row uses ${\approx}1{,}490$ iterations and the
  25B-token subset-selection rows use ${\approx}745$ iterations, so compute is
  matched within the 25B-token block.
}
\label{tab:midtrain-config}
\begin{tabular}{ll|ll}
\toprule
\textbf{Field} & \textbf{Value} & \textbf{Field} & \textbf{Value} \\
\midrule
\multicolumn{4}{c}{\textit{Architecture (Qwen2.5-Coder-14B)}} \\
\midrule
Layers              & 48                & Attention heads       & 40 \\
Hidden size         & 5{,}120           & KV heads (GQA)        & 8 \\
FFN hidden size     & 13{,}824          & Max position emb.\    & 131{,}072 \\
Activation          & SwiGLU            & Positional encoding   & RoPE (base $=10^{6}$) \\
Normalization       & RMSNorm           & RMSNorm $\varepsilon$ & $10^{-5}$ \\
Embedding tying     & untied            & Linear bias           & off (QKV bias on) \\
Attn / hidden dropout & 0 / 0           & Vocab divisor         & 64 \\
\midrule
\multicolumn{4}{c}{\textit{Precision and Parallelism}} \\
\midrule
Precision           & BF16              & Sequence parallel     & on \\
Sequence length     & 131{,}072 (128K)  & TP comm.\ overlap     & on \\
Seed                & 42                & Distributed optimizer & on \\
TP / PP / CP / EP   & 8 / 1 / 4 / 1     & Overlap grad / param  & on \\
\midrule
\multicolumn{4}{c}{\textit{Optimization}} \\
\midrule
Optimizer           & AdamW             & Weight decay          & 0.1 \\
$(\beta_1, \beta_2, \varepsilon)$ & $(0.9, 0.95, 10^{-8})$
                                        & Gradient clipping     & 1.0 \\
LR schedule         & cosine            & Init.\ std.\          & 0.02 \\
LR range            & $5\!\times\!10^{-5} \to 1\!\times\!10^{-5}$
                                        & LR warmup iters       & 5 \\
Micro / global batch & 1 / 256          & EOD masking           & on \\
\midrule
\multicolumn{4}{c}{\textit{Checkpointing and Reproducibility}} \\
\midrule
Save format         & torch\_dist (async) & Persistent save & every 1{,}000 iters \\
Non-persistent save & every 500 iters   & Restart load          & no optimizer, no RNG \\
Deterministic mode  & on                & NCCL algorithm        & \^{}NVLS \\
TE nondet.\ algos   & disabled          & cuBLAS workspace      & :4096:8 \\
\bottomrule
\end{tabular}
\vspace{-5pt}
\end{table*}

\paragraph{Token budget and iteration count.}
The total token budget is set by the total-tokens launch argument and
converted to iterations as
$\text{train-iters} = \lfloor \text{tokens} / (256 \cdot 131{,}072) \rfloor$,
so one optimizer step consumes $256 \times 131072 = 33{,}554{,}432$ tokens.
The full-corpus row of Table~\ref{tab:main} uses a 50B-token budget
(${\approx}1{,}490$ iterations). The 25B-token subset-selection rows
(MIRA variants, DataMan, DSIR, PPL, Random) all use ${\approx}745$
iterations, so iteration count is matched within that block.
When no resumable checkpoint exists, the finetune flag is appended
automatically, ensuring the run starts from Qwen2.5-Coder-14B weights
with a fresh optimizer state. For every condition, the final-iteration
checkpoint (iter\_0000745) is the only artifact handed off to
the SFT stage of \S\ref{app:sft-config}.


\subsection{SFT Configuration}
\label{app:sft-config}

All mid-training checkpoints are post-trained under the same SFT recipe
before evaluation, so the post-SFT numbers in Table~\ref{tab:main}
isolate the mid-training corpus from the supervised stage. The SFT
recipe is full fine-tuning of all parameters in Megatron-LM with
sequence packing enabled. Table~\ref{tab:sft-config}
records the recipe, and Table~\ref{tab:sft-vs-mt} summarizes the
settings that differ from the mid-training stage.

\begin{table*}[!t]
\small
\centering
\caption{
  SFT recipe applied identically to every mid-training checkpoint
  (and to the Qwen2.5-Coder-14B base weights for the SFT-only row of
  Table~\ref{tab:main}). The base architecture (layers, hidden size,
  GQA, RoPE base, SwiGLU, etc.) is reused from Table~\ref{tab:midtrain-config}
  and omitted here.
}
\label{tab:sft-config}
\begin{tabular}{ll|ll}
\toprule
\textbf{Field} & \textbf{Value} & \textbf{Field} & \textbf{Value} \\
\midrule
\multicolumn{4}{c}{\textit{Data and Initialization}} \\
\midrule
Init.\ weights      & mid-training final ckpt           & Tokenizer / prompt format & SFTTokenizer / qwen2 \\
Data                & 400k\_subset.jsonl (400K)& Sequence packing      & on \\
Epochs              & 2                                 & Train iters           & $2 \lfloor P / 512 \rfloor$, $P{=}$ packed samples \\
Restart load        & no optim, no RNG (finetune flag) & Save interval    & every 500 iters \\
\midrule
\multicolumn{4}{c}{\textit{Precision and Parallelism}} \\
\midrule
Precision           & BF16                              & Sequence parallel     & on \\
Sequence length     & 128{,}768                         & Distributed optimizer & on \\
Seed                & 42                                & Activation recompute  & full, uniform, 1 layer \\
TP / PP / CP / EP   & 8 / 1 / 1 / 1                     & RMSNorm $\varepsilon$ & $10^{-6}$ \\
\midrule
\multicolumn{4}{c}{\textit{Optimization}} \\
\midrule
Optimizer           & AdamW                             & Weight decay          & 0.01 \\
$(\beta_1, \beta_2, \varepsilon)$ & $(0.9, 0.95, 10^{-8})$
                                                        & Gradient clipping     & 1.0 \\
LR schedule         & cosine                            & Init.\ std.\          & 0.02 \\
LR range            & $5\!\times\!10^{-5} \to 1\!\times\!10^{-5}$
                                                        & LR warmup iters       & $\lfloor \text{TRAIN\_ITER} / 10 \rfloor$ \\
Micro / global batch & 1 / 512                          & EOD masking           & on \\
\bottomrule
\end{tabular}
\vspace{-5pt}
\end{table*}

\begin{table*}[!t]
\small
\centering
\caption{
  Settings that differ between mid-training and SFT.
  All other hyperparameters (optimizer, $\beta$s, $\varepsilon$,
  LR schedule shape and endpoints, gradient clipping, seed, and base
  architecture) are identical across the two stages.
}
\label{tab:sft-vs-mt}
\begin{tabular}{lll}
\toprule
\textbf{Field}         & \textbf{Mid-training}                 & \textbf{SFT} \\
\midrule
Global batch           & 256                                   & 512 \\
Sequence length        & 131{,}072                             & 128{,}768 \\
Context parallel       & 4                                     & 1 \\
Weight decay           & 0.1                                   & 0.01 \\
RMSNorm $\varepsilon$  & $10^{-5}$                             & $10^{-6}$ \\
LR warmup              & 5 iters (fixed)                       & $\lfloor \text{TRAIN\_ITER} / 10 \rfloor$ \\
Activation recompute   & off                                   & full, uniform, 1 layer \\
Sequence packing       & off                                   & on \\
Tokenizer mode         & HuggingFaceTokenizer         & SFTTokenizer (qwen2) \\
Restart behavior       & finetune from base         & finetune, no optim, no RNG \\
\bottomrule
\end{tabular}
\vspace{-5pt}
\end{table*}

\paragraph{Sequence packing and effective batch.}
With sequence packing enabled, multiple short SFT samples are
concatenated into single 128K-token sequences during preprocessing.
The resulting packed\_samples count $P$ and a packing ratio
are read from the preprocessing-info file at launch time.
TRAIN\_ITER is then set to $2 \lfloor P / 512 \rfloor$,
i.e.\ two epochs over the packed dataset, and the warmup is one tenth
of that. The 400K original SFT records therefore translate to a much
smaller optimizer-step count than a naive $400{,}000 / 512$ would
suggest, and this packed iteration count is what governs the cosine
LR decay.

\paragraph{Loss target and masking.}
Loss is computed on the assistant response tokens only. System,
user, and tool-result tokens are masked via the SFT tokenizer's prompt
format (qwen2) plus the EOD mask-loss flag, so
neither the instruction nor the document boundary contributes to the
gradient.

\paragraph{Held constants across conditions.}
Because the same SFT recipe is applied to every mid-trained checkpoint
in Table~\ref{tab:main} (Random, DSIR, PPL, DataMan, MIRA-Global,
MIRA-Group, MIRA-Source, Raw-Mixture, and the SFT-only baseline that
skips mid-training), differences in the post-SFT macro-average reflect
only the mid-training data selection strategy. The SFT-only row of
Table~\ref{tab:main} uses the same recipe but initializes from the
Qwen2.5-Coder-14B base weights directly, establishing the
no-mid-training baseline.


\subsection{Teacher and Student Scorers}
\label{app:scorers}

The scoring stack of \S\ref{sec:rubric-discovery} and \S\ref{sec:distillation}
has two components: a single frontier teacher used for anchored
labeling across all groups, and additionally for free-dim rubric
discovery on the Text and Agent tracks, and a family of group-specific
student scorers distilled from the teacher's anchored labels. At the
experiment-time granularity of \S\ref{app:data}, this gives one
teacher and 5 student scorers, one per experiment group.

\paragraph{Teacher.}
We use Kimi-K2.6 as the frontier judge for both Phase-1 (free-dim) and
Phase-2 (anchored) calls. The teacher receives the prompts of
\S\ref{app:prompts}. It is invoked with temperature 0 and a 38{,}000-token
prompt budget (roughly 128{,}000 user characters after chat-template
overhead). Each Phase-1 record (all five groups) is scored on the
teacher-proposed 15 dimensions. Each Phase-2 record (all five groups)
is rescored on the 15 anchor dimensions fixed for its group. All five
groups derive their 15 anchors via K-means $k{=}15$ on Phase-1
judgment-point embeddings (\S\ref{app:prompt-phase2}), so the QA,
Text, and Agent tracks share an identical two-stage
discovery-then-scoring pipeline.

\paragraph{Student scorers.}
We distill 5 group-specific students from the Phase-2 anchored teacher
labels, one per experiment-time group: qa\_math\_reasoning,
qa\_code\_general, text\_code\_doc, agent\_toolcall, agent\_swe. Each
student is fine-tuned to emit a (name, score, reason) triple for each
of the 15 anchor dimensions of its group, parsed by the same
[A1]..[A15] regex used on teacher output. Total training data
across the 5 students is roughly 2M teacher-scored records. The
held-out validation split (Table~\ref{tab:student-eval}) is used
exclusively for the reliability diagnostics of \S\ref{app:reliability}.

\paragraph{Dimension labeling.}
For uniformity across tracks, every group is described in the
remainder of the appendix as having 15 slots labeled \textbf{A1..A15}.
We keep this unified notation for presentation and flag cells beyond A15
explicitly in Table~\ref{tab:reliability-mask}.

\begin{table}[!t]
\small
\centering
\caption{
  Held-out validation split sizes per source for the Phase-2
  teacher/student agreement evaluation (\S\ref{app:reliability}).
  ``$n_{\text{val}}$'' is the number of records where both teacher and
  student outputs parsed successfully. ``Avg MAE'' is the overall
  teacher/student mean absolute error averaged across A1..A15.
  Sources are listed under their experiment-time group of \S\ref{app:data}.
}
\label{tab:student-eval}
\begin{tabular}{lrr}
\toprule
\textbf{Source} & $\mathbf{n_{\text{val}}}$ & \textbf{Avg MAE} \\
\midrule
\multicolumn{3}{c}{\textit{qa\_math\_reasoning (5 sources)}} \\
\midrule
math\_theorem               & 23{,}590 & 0.51 \\
math\_code               & 23{,}734 & 0.49 \\
math\_reason\_a                    & 22{,}514 & 0.58 \\
math\_reason\_b         & 21{,}562 & 0.46 \\
math\_reason\_c           & 23{,}623 & 0.34 \\
\midrule
\multicolumn{3}{c}{\textit{qa\_code\_general (5 sources)}} \\
\midrule
code\_qa\_a & 15{,}685 & 0.45 \\
code\_qa\_b    & 14{,}189 & 0.51 \\
code\_qa\_c          & 14{,}246 & 0.52 \\
code\_qa\_d            & 11{,}435 & 0.46 \\
general\_qa           & 18{,}279 & 0.51 \\
\midrule
\multicolumn{3}{c}{\textit{text\_code\_doc (3 sources)}} \\
\midrule
algo\_text         & 50{,}163 & 0.72 \\
pr\_text              & 33{,}238 & 0.55 \\
code\_contest\_qa${}^{*}$   & 10{,}887 & 0.68 \\
\midrule
\multicolumn{3}{c}{\textit{agent\_toolcall (5 sources)}} \\
\midrule
cli\_agent\_a            &  4{,}319 & 0.85 \\
cli\_agent\_b             &  2{,}119 & 0.76 \\
funccall\_agent\_a        &  3{,}776 & 0.52 \\
funccall\_agent\_b               &  3{,}009 & 1.00 \\
funccall\_agent\_c                 &  3{,}550 & 0.52 \\
\midrule
\multicolumn{3}{c}{\textit{agent\_swe (3 sources)}} \\
\midrule
agent\_a       &  1{,}797 & 0.55 \\
agent\_b            &  2{,}506 & 0.60 \\
agent\_c                &  1{,}145 & 0.43 \\
\bottomrule
\end{tabular}
\vspace{-5pt}
\end{table}

\subsection{Student Scorer Training Configuration}
\label{app:student-train}

Each of the 5 group-specific student scorers introduced in
\S\ref{app:scorers} is obtained by full-parameter supervised
fine-tuning of a shared mixture-of-experts (MoE) base model on the
group's Phase-2 anchored teacher labels. All five students share a
single optimization recipe and differ only in their training data path
and the per-group sequence-length budget. Group-level
\texttt{cutoff\_len} and batch settings track the length distribution
of the underlying source mix rather than any deliberate per-group
tuning.

\paragraph{Base model.}
We use a 35B-parameter MoE decoder with $\approx 3$B active parameters
per token (Qwen3.5-35B-A3B-Base class). Full-parameter SFT is used
rather than parameter-efficient adaptation so that the student can
learn the structured [A1]..[A15] output schema directly rather than
overlaying it on a frozen instruction prior. The same base weights are
used for every group, so cross-group differences in student behavior
reflect only the group's anchored teacher labels.

\paragraph{Shared recipe.}
Table~\ref{tab:student-config} lists the optimization and runtime
settings shared by all 5 students. Sequence packing is enabled to
amortize attention compute over the long-tailed sample-length
distribution within each group. Gradient checkpointing with
non-reentrant autograd bounds activation memory on the
longer-context groups. ZeRO-3 (no offload) with BF16 plus
flash-attention v2 is used as a single uniform parallelism setting
across all groups, so cross-group comparisons are not confounded by
parallel-config differences.

\begin{table}[!t]
\small
\centering
\caption{
  Shared SFT recipe for all 5 student scorers. Group-specific
  \texttt{cutoff\_len} is set to $\approx p_{95}$ of the packed
  sequence-length distribution measured on a 5K-record tokenization
  sample: 24{,}576 for qa\_math\_reasoning (chain-of-thought-heavy)
  and 16{,}384 for the remaining four groups. Per-device micro-batch
  is 1 with gradient accumulation 8 throughout, holding the
  optimizer-step batch constant across groups.
}
\label{tab:student-config}
\resizebox{\columnwidth}{!}{%
\begin{tabular}{ll}
\toprule
\textbf{Field} & \textbf{Value} \\
\midrule
Base model            & 35B MoE, $\approx 3$B active \\
Finetuning type       & full parameters \\
Precision             & BF16 + flash-attention v2 \\
Parallelism           & DeepSpeed ZeRO-3 (no offload) \\
Optimizer             & AdamW \\
LR schedule           & cosine, warmup ratio 0.05 \\
Peak / final LR       & $5\!\times\!10^{-6}$ \\
Epochs                & 2 \\
Sequence packing      & on \\
Gradient checkpointing & on (non-reentrant) \\
\texttt{cutoff\_len}  & 24K / 16K (see caption) \\
Per-device micro batch & 1 \\
Gradient accumulation & 8 \\
Eval / save interval  & every 500 steps (keep 3) \\
\bottomrule
\end{tabular}}
\vspace{-5pt}
\end{table}

\paragraph{Compute.}
End-to-end student-scorer SFT consumes approximately
$200 / 0.8 \approx 250$ H800 GPU-hours aggregated across all 5
students, or roughly $50$ H800 GPU-hours per student. The numerator
$200$ is the ideal-throughput estimate derived from $6 \cdot
N_{\text{active}} \cdot N_{\text{tok}}$ training FLOPs, with
$N_{\text{active}} \approx 3 \times 10^{9}$ active parameters per token
on the MoE backbone and $N_{\text{tok}} \approx 4 \times 10^{10}$
aggregate token-passes (the $\approx 2$M-record Phase-2 corpus of
\S\ref{app:scorers} trained for 2 epochs at $\approx 10$K average
packed sequence length), divided by H800 BF16 peak throughput
($\approx 989$ TFLOPS). The $0.8$ divisor is the same empirical
utilization factor used in \S\ref{app:midtrain-config}. This places
student distillation at less than $4\%$ of the per-condition
mid-training cost of \S\ref{app:midtrain-config}
(${\approx} 7{,}680$ H800 GPU-hours), so adding rubric discovery and
student distillation does not materially change the overall compute
budget.

\subsection{Reliability Diagnostics}
\label{app:reliability}

The reliability mechanism of \S\ref{sec:reliability} is operationalized
by computing per-(source, slot) MAE between teacher and student on the
held-out validation split of Table~\ref{tab:student-eval}, then masking
slots whose disagreement exceeds a threshold. We use
\textbf{$\text{MAE} \geq 1$} (on the 0 to 10 scoring scale, equivalent
to roughly one 5-point training-utility band) as the masking rule.
Cells with $\text{MAE} < 1$ contribute to the trimmed-mean
aggregation. Cells at or above the threshold are zeroed out and the
remaining slots are averaged.

\paragraph{Masked cells.}
Table~\ref{tab:reliability-mask} lists the (source, slot) cells flagged
by the $\text{MAE} \geq 1$ rule across the 21 mid-training sources.
Slot indices use the unified A1..A15 notation. For
code\_contest\_qa the index range extends to A16..A20
(qa-with-think variant, \S\ref{app:scorers}). The right-most column
summarizes per-source disagreement intensity by reporting the
largest per-slot MAE.

\begin{table*}[!t]
\small
\centering
\caption{
  Slot semantics are anchor-list-specific: identical slot indices on
  different sources do not necessarily denote the same dimension
  (e.g.\ A8 in qa\_math\_reasoning maps to ``Technical Precision'',
  while A8 in agent\_toolcall maps to ``Action Efficiency''). ``Max
  MAE'' is the largest per-slot MAE on that source.
}
\label{tab:reliability-mask}
\begin{tabular}{l p{6cm} r}
\toprule
\textbf{Source} & \textbf{Masked slots} & \textbf{Max MAE} \\
\midrule
\multicolumn{3}{c}{\textit{qa\_math\_reasoning (3 of 5 sources)}} \\
\midrule
math\_reason\_a                   & A1, A8        & 0.907 \\
math\_theorem              & A1, A8        & 0.878 \\
math\_code              & A8            & 0.852 \\
\midrule
\multicolumn{3}{c}{\textit{qa\_code\_general (1 of 5 sources)}} \\
\midrule
code\_qa\_c         & A8            & 0.806 \\
\midrule
\multicolumn{3}{c}{\textit{text\_code\_doc (3 of 3 sources)}} \\
\midrule
algo\_text        & A2, A4, A5, A13, A14 & 0.959 \\
pr\_text             & A15           & 1.235 \\
code\_contest\_qa${}^{*}$  & A8, A13, A16  & 0.918 \\
\midrule
\multicolumn{3}{c}{\textit{agent\_toolcall (3 of 5 sources)}} \\
\midrule
cli\_agent\_a           & A1, A2, A5, A10, A13 & 1.463 \\
cli\_agent\_b            & A2, A3, A5, A6 & 2.034 \\
funccall\_agent\_b              & A1, A5, A8, A9, A11, A12, A14, A15 & 1.397 \\
\midrule
\multicolumn{3}{c}{\textit{agent\_swe (2 of 3 sources)}} \\
\midrule
agent\_a      & A3            & 0.908 \\
agent\_b           & A3, A10, A11, A13 & 1.223 \\
\bottomrule
\end{tabular}
\vspace{-5pt}
\end{table*}

\paragraph{Clean sources.}
Of the 21 mid-training sources, 9 have no flagged slot under the
$\text{MAE} \geq 1$ rule and contribute their full 15 dimensions to
the trimmed-mean aggregation:
\emph{qa\_math\_reasoning (2 of 5)}: math\_reason\_b, math\_reason\_c.
\emph{qa\_code\_general (4 of 5)}: code\_qa\_a,
code\_qa\_b, code\_qa\_d, general\_qa.
\emph{text\_code\_doc (0 of 3)}: none (every text\_code\_doc source
has at least one flagged cell, consistent with the heterogeneity of
the long-form / extended-schema content this group covers).
\emph{agent\_toolcall (2 of 5)}: funccall\_agent\_a, funccall\_agent\_c.
\emph{agent\_swe (1 of 3)}: agent\_c.

\paragraph{Aggregate mask coverage.}
The mask rule flags 37 source/slot cells across the 21 mid-training
sources: 5 in qa\_math\_reasoning, 1 in qa\_code\_general, 9 in
text\_code\_doc , 17 in agent\_toolcall, and 5 in agent\_swe.
Relative to a $21 \times 15 = 315$-cell A1..A15 budget , the masked fraction is
$37 / 315 \approx 12\%$. The flagged cells concentrate on a small
number of axes. ``Technical Precision'' is the most frequently masked dimension on the QA-format
sources, flagging in 5 of the 11 QA-format records (math\_reason\_a,
math\_theorem, math\_code, code\_qa\_c, and code\_contest\_qa). On the
text\_code\_doc side, format-compliance axes account for the worst-MAE
cells, namely ``Structural Organization'' (A13) on algo\_text and
``Output Format Adherence'' (A15) on pr\_text, reflecting a known
student weakness on structured output. In Agent, ``Code Reference
Accuracy'' (A2 in agent\_toolcall's anchor list) is flagged on every
CLI source (cli\_agent\_a, cli\_agent\_b) because the student cannot
ground tool-call references back to filesystem state without
inspection. These patterns motivate the \emph{post-hoc} masking
strategy of \S\ref{sec:reliability}: rather than removing or
relabeling the affected dimensions inside the student prompt (which
would change the joint distribution of the remaining 14 slots), we
leave the prompt fixed and zero the affected cells only at
aggregation time.

\paragraph{Worst-case source.}
funccall\_agent\_b (agent\_toolcall) is the worst-case source in the
mid-training pool with 8 of 15 slots flagged, consistent with its
overall MAE of 1.00 and concentrated on tool-call argument and
safety-scope axes. We retain it because the surviving 7 slots still
provide a usable composite. Under trimmed-mean aggregation,
funccall\_agent\_b's overall score is computed over those 7 surviving slots, so
its score is effectively determined by a narrower subset of axes than
other agent sources.


\section{Broader Impact}
\label{sec:broader-impact}

MIRA is intended to improve the efficiency and transparency of mid-training data
selection by making quality criteria explicit at the source-group level. This
can reduce unnecessary training compute by selecting smaller but more useful
subsets, and can help practitioners audit why particular data sources are
retained or removed. At the same time, learned filtering systems can amplify the
preferences and blind spots of the teacher model used to construct rubrics. If
the teacher systematically undervalues certain programming styles, languages,
domains, or user communities, the resulting scorer may reproduce those biases at
scale. We therefore view MIRA as a tool for structured data auditing rather than
as a replacement for human review, contamination checks, license compliance, or
safety evaluation.

Because our experiments focus on code-oriented mid-training, the most direct
impacts are on code generation and software-engineering assistants. Better data
selection may improve developer productivity and reduce inefficient training
runs, but stronger coding models can also lower the barrier to generating
vulnerable code, automating misuse workflows, or overfitting to benchmark-like
programming tasks. Deployments of MIRA-filtered models should therefore be
paired with security evaluation, provenance tracking, and downstream monitoring
appropriate to the application context.

\section{Prompts}
\label{app:prompts}

This section reproduces the prompt templates MIRA sends to the frontier
teacher at both rubric-discovery time, Phase-1 (\S\ref{sec:rubric-discovery},
Figure~\ref{lst:prompt-freedim}), and anchored labeling time,
Phase-2 (\S\ref{sec:distillation}, Figure~\ref{lst:prompt-anchored}). Every
prompt elicits \textbf{15 quality dimensions} per record, emitted as
parseable [A1]..[A15] lines. Phase-1 lets the teacher freely name those
15 dimensions. Phase-2 fixes them to the anchors of the record's source
group, discovered by K-means clustering of Phase-1 judgment points.
For readability, the bullet and double hyphen characters shown below are
ASCII renderings of the Unicode characters used in the live prompts.

\subsection{Phase-1: Free-Dimension Rubric Discovery}
\label{app:prompt-phase1}

The free-dim prompt asks the teacher to (i) propose 15 quality dimensions
it considers most relevant to the sample, (ii) score each on a 0 to 10
integer scale anchored to five training-utility bands, and (iii) emit
[A$k$]-prefixed lines that the parser of
\S\ref{app:prompt-parse} can recover. The full free-dim template is
shown in Figure~\ref{lst:prompt-freedim}. The same prompt is applied
uniformly across the Text, Agent, and QA tracks, with the teacher free
to choose whichever 15 dimensions it judges most relevant to the sample
at hand.

\begin{figure*}[!t]
\centering
\begin{tcolorbox}[
    colback=white!95!gray,
    colframe=black,
    width=\linewidth,
    arc=4mm,
    boxrule=0.5mm
]
\raggedright
\textbf{\large Phase-1 Free-Dimension Prompt}

\vspace{0.5em}
\begin{lstlisting}[style=trace, frame=none, aboveskip=0pt, belowskip=0pt, xleftmargin=0pt]
[SYSTEM]
You are a world-class data quality evaluator for AI training datasets.
Below you'll see a piece of training data (may include long system
prompts, user requests, assistant responses, agent traces, technical
docs, or any mixture). Your job is to evaluate it across 15 dimensions
that YOU select as most relevant for this specific kind of content.
Score each dimension on a 0-10 integer scale using the training-utility
bands below.

Be strict. Most samples should fall in middle bands, reserving 9-10 for
truly exceptional samples and 0-1 for completely broken samples. Do NOT
default to the top.

Training-utility bands:
  0-1  : completely untrainable -- actively harmful if included
  2-3  : not trainable -- major defects
  4-5  : passable -- usable only in bulk, low signal
  6-7  : trainable -- meets quality bar for inclusion
  8-10 : strongly recommended -- high-value exemplar
         (10 = near-perfect, rare)

[USER]
Evaluate the following training sample. Choose 15 dimensions YOU consider
most critical for judging this kind of data. Possible angles include
(NOT limited to): factual correctness, completeness, clarity, structure,
reasoning quality, practical applicability, technical depth, format
consistency, instruction-following, safety, communication quality,
signal-to-noise ratio, training utility, diversity, domain expertise, etc.

---
# Source: {source}
# Domain: {domain}

# Content
{text}
---

## Scoring scale (0-10 integer)
Anchor each score to a band:
  [0-1]  completely untrainable -- actively harmful if included
  [2-3]  not trainable -- major defects
  [4-5]  passable -- usable only in bulk, low signal
  [6-7]  trainable -- meets quality bar for inclusion
  [8-10] strongly recommended -- high-value exemplar
         (10 near-perfect, rare)

Calibration rules:
- A score of 10 means "could not be meaningfully improved". Use sparingly.
- If a dimension has any notable weakness, cap it at 7.
- If a dimension is merely acceptable with no standout strength, score
  4-5, not 6+.
- Use the full range. Do not cluster every dimension near the top.

## Part A -- 15 Self-defined Quality Dimensions
For each of YOUR 15 chosen dimensions:
1. Name it (specific, not vague)
2. Give 0-10 integer score
3. One-sentence justification

Format strictly:
[A1]  <Dimension Name>: <score>/10 -- <justification>
[A2]  <Dimension Name>: <score>/10 -- <justification>
...
[A15] <Dimension Name>: <score>/10 -- <justification>

## Part B -- Summary
[Overall Score]: <average of A1-A15, rounded to 1 decimal>/10
[Training Recommendation]: <untrainable | not_trainable | passable
                          | trainable | strongly_recommended>
[Domain Tag]: <one short label, e.g. agent_trace / cli_doc / sql_qa /
              pr_fix / etc.>
[Brief Assessment]: <2-3 sentences summarizing key strengths and weaknesses>
\end{lstlisting}
\end{tcolorbox}
\caption{Phase-1 free-dim prompt, applied uniformly across Text, Agent,
and QA tracks.}
\label{lst:prompt-freedim}
\end{figure*}
\subsection{Phase-2: Anchored Scoring}
\label{app:prompt-phase2}
After fixing 15 anchor dimensions per source group via K-means
clustering of Phase-1 judgment points (see \S\ref{sec:rubric-discovery}),
Phase-2 re-prompts the teacher with those anchors as fixed
[A1]..[A15] slots (Figure~\ref{lst:prompt-anchored}). The same template
is applied uniformly across all tracks. Only the rendered \{dim\_lines\}
block and the in-context calibration block (\S\ref{app:prompt-calib})
vary across source groups.

\begin{figure*}[!t]
\centering
\begin{tcolorbox}[
    colback=white!95!gray,
    colframe=black,
    width=\linewidth,
    arc=4mm,
    boxrule=0.5mm
]
\raggedright
\textbf{\large Phase-2 Anchored Scoring Prompt}

\vspace{0.5em}
\begin{lstlisting}[style=trace, frame=none, aboveskip=0pt, belowskip=0pt, xleftmargin=0pt]
[SYSTEM]
You are a world-class data quality evaluator for AI training datasets.
Score each dimension strictly on a 0-10 integer scale using the
training-utility bands below. Be strict. Most samples should fall in
the middle bands, reserving 9-10 for truly exceptional samples and 0-1
for completely broken ones. Do NOT default to the top of the scale.

Training-utility bands (apply to every dimension and the overall score):
  0-1  : completely untrainable -- actively harmful if included
  2-3  : not trainable -- major defects
  4-5  : passable -- usable only in bulk, low signal
  6-7  : trainable -- meets quality bar for inclusion
  8-10 : strongly recommended -- high-value exemplar
         (10 = near-perfect, rare)

(... calibration references block, see lst:prompt-calib ...)

[USER]
Evaluate the following training sample on exactly 15 fixed quality
dimensions defined for this data group.

---
# Source: {source}
# Domain: {domain}

# Content
{content}
---

## Scoring scale (0-10 integer, apply to every dimension)
(... identical bands and calibration rules as Phase-1 ...)

## Part A -- 15 Fixed Dimensions (Group {group_id})
Score each dimension 0-10 (integer). One-sentence justification required.

[A1]  {anchor_1_name}:  <score>/10 -- <justification>
[A2]  {anchor_2_name}:  <score>/10 -- <justification>
...
[A15] {anchor_15_name}: <score>/10 -- <justification>

## Part B -- Summary
[Overall Score]: <average of A1-A15, rounded to 1 decimal>/10
[Training Recommendation]: <untrainable | not_trainable | passable
                          | trainable | strongly_recommended>
[Domain Tag]: <one short label>
[Brief Assessment]: <2-3 sentences summarizing key strengths and weaknesses>
\end{lstlisting}
\end{tcolorbox}
\caption{Phase-2 anchored scoring prompt, applied uniformly across all
tracks.}
\label{lst:prompt-anchored}
\end{figure*}
\subsection{In-Context Calibration Block}
\label{app:prompt-calib}

To stabilize Phase-2 scoring across millions of records, the SYSTEM
message is suffixed with a \emph{calibration references} block listing
the top-12 anchors of the record's group (the same set whose
anchor\_point centroids name the [A1]..[A15] slots,
truncated to 12 for prompt-length budget). Each reference contributes a
(name, score, one-sentence reason) triple drawn from the cluster's
representative judgment point, so the teacher sees concrete worked
examples of how severity has been applied to comparable defects.

\subsection{Response Parsing}
\label{app:prompt-parse}

Teacher responses are parsed with a single regex that tolerates Unicode
colon and dash variants. A record is kept only if at least 12 of the 15
dimension lines parse successfully. Surviving records contribute their
$(\text{name}, \text{score}, \text{reason})$ triples to either Phase-1
clustering (free-dim) or the Phase-2 train/validation splits (anchored).

\end{document}